\documentclass[
]{llncs}

\usepackage{relsize,etoolbox}%
\usepackage[utf8]{inputenc}
\usepackage{amsmath}
\usepackage{graphicx}
\usepackage{amsfonts}
\usepackage{amssymb}
\usepackage{url}

\AtBeginEnvironment{quote}{\smaller}

\begin{document}

{\let\thefootnote\relax\footnotetext{Copyright \textcopyright\ 2020 for this paper by its authors. Use permitted under Creative Commons License Attribution 4.0 International (CC BY 4.0). CLEF 2020, 22-25 September 2020, Thessaloniki, Greece.}}



\title{Accenture at CheckThat! 2020: If you say so: Post-hoc fact-checking of claims using transformer-based models}


\author{Evan Williams\inst{1}\orcidID{0000-0002-0534-9450} \and Paul Rodrigues\inst{1}\inst{2}\orcidID{0000-0002-2151-636X} \and Valerie Novak\inst{2}\orcidID{0000-0001-8317-0993}}
\institute{
Accenture,
  800 N. Glebe Rd., Arlington, 22209, USA\\
\url{e.m.williams@accenture.com}\\
\url{paul.rodrigues@accenture.com}
\and
University of Maryland,
  College Park, MD, USA\\
\and
\url{vnovak@umd.edu}
}

\maketitle

\begin{abstract}
  We introduce the strategies used by the Accenture Team for the CLEF2020 CheckThat! Lab, Task 1, on English and Arabic.  This shared task evaluated whether a claim in social media text should be professionally fact checked.  To a journalist, a statement presented as fact, which would be of interest to a large audience, requires professional fact-checking before dissemination.  We utilized BERT and RoBERTa models to identify claims in social media text  a professional fact-checker should review, and rank these in priority order for the fact-checker. For the English challenge, we fine-tuned a RoBERTa model and added an extra mean pooling layer and a dropout layer to enhance generalizability to unseen text.  For the Arabic task, we fine-tuned Arabic-language BERT models and demonstrate the use of back-translation to amplify the minority class and balance the dataset. The work presented here was scored 1st place in the English track, and 1st, 2nd, 3rd, and 4th place in the Arabic track.
\end{abstract}



\section{Introduction}

Natural Language Processing (NLP) has been driving Artificial Intelligence research since the 1950s, but recently increased in distinction due to the quantity of text that can be utilized as well as new techniques to extract even more value from text. In 2018, a surge of research produced deep learning architectures in NLP which beat state of the art on a multitude of tasks, such as sentiment analysis, question answering, and semantic similarity, in a variety of languages.  Since the innovation of ULMFit \cite{howard:2018}, numerous new architectures have been introduced, such as ELMo \cite{peters:2018}, BERT \cite{devlin2018bert}, ERNIE \cite{sun:2019}, RoBERTa \cite{roberta}, GPT-2 \cite{radford:2019}, GPT-3 \cite{brown:2020}, and others, yielding breakthrough innovations and increased performance, nearly month after month.  These  architectures require massive amounts of training data, which can be expensive to train on high-performance computing clusters \cite{sharir:2020}.  However, they facilitate the practice of transfer learning. A base model trained on a large amount of general text data can then be fine-tuned, or customized for a specific problem and domain/genre, using text with far less annotated data than previous systems required.  This use of transfer learning allows us to effectively craft custom cutting-edge models to solve a wide range of classification problems.

While these architectures are often utilized to improve NLP tasks, the application of transformer-based transfer learning approaches are less often demonstrated as components in decision-support systems which aid the workflow of subject matter experts. We do see these technologies being used in the medical field (e.g. \cite{rasmy2020medbert}), and anticipate there will be many more  applications coming.  The CheckThat! Lab poses one such application, which could reduce information burden in the workflow of a journalist.    


\subsection{CheckThat! Lab}


We participated in Task 1 of the 2020 CheckThat! challenge. \cite{clef-checkthat-lncs:2020} Organizers distributed  collections of tweets in English and in Arabic for training, annotated for topic group, whether the tweet was a claim, and whether the tweet was check-worthy, along with Twitter provided meta-data. \cite{clef-checkthat-en:2020,clef-checkthat-ar:2020} Participants in the challenge utilized this data to train a model that could receive a list of novel tweets, classify each for check-worthiness, and rank the group of tweets by how check-worthy they were. Evaluation of the model was performed on a second test dataset provided for each language. These test datasets were held back by the organizers until shortly before the competition end time.  Organizers provided this dataset unlabeled, and participants provided the labels and ranking to the organizers.  Organizers evaluated the ranking produced by participating groups to a withheld labeled and ranked list.  Participants were permitted to submit one primary run and up to 3 contrasting runs.  The official metric for Arabic was Precision @ 30 (P@30).  Precision @ \textit{k} is the number of relevant results in the top \textit{k} claims in the ranked list.  The official metric for English was Mean Average Precision (mAP), or the mean of the average precision scores for each of the claims.

\subsubsection{Provided Data}
Tweets were collected by CheckThat! organizers using keyword watchlists, consisting of usernames, hashtags, or key words, designed around a variety of topic areas.

For English, one topic was provided related to COVID-19, and filtered for tweets that mentioned \#COVID19, \#CoronavirusOutbreak, \#Coronavirus, \#Corona, \#CoronaAlert, \#CoronaOutbreak, corona, and COVID-19.  This topic was the same in train, test, and the evaluation set.

For Arabic, the training data included three topics--Protests in Lebanon, Emirati cleric Wassim Youssef, as well as Turkey's intervention in Syria.  Testing data included topics such as Deal of the Century, The Houthis in Yemen, COVID-19, Feminists, Events in Libya, The group of resident non-citizens in Kuwait, Algeria, as well as Boycotting Countries \& Promoting Rumors against Qatar.  We note that the topics provided between train and test datasets differ, with no overlap.

The topic word lists were used by the organizers to collect posts on Twitter.  Annotators were presented these posts and were asked to evaluate each for check-worthiness. Check-worthiness was defined as ``a tweet that includes a claim that is of interest to a large audience (especially journalists), might have a harmful effect, etc." \cite{checkthat2020} Tweets were assigned check-worthiness labels after review by two annotators as well as a review by a third expert annotator. Check-worthiness was evaluated on the following three criteria \cite{CLEF_LNCS:20}:

\begin{itemize}
\item Do you think the claim in the tweet is of interest to the public?
\item To what extent do you think the claim can negatively affect the reputation
of an entity, country, etc.?
\item Do you think journalists will be interested in covering the spread of the claim
or the information discussed by the claim?
\end{itemize}

In examining the labeled training data, we confirmed nuanced differences between tweets that were check-worthy and tweets that were not. For example, the tweet below, which was taken from the English task development data, initially appears to be peddling a false COVID-19 claim. However, the rest of the tweet makes it clear that the author is joking, which is presumably why this tweet was not labeled as being check-worthy.

\begin{quote}
\centering
"ALERT!!!!!! The corona virus can be spread through money. If you have any money at home, put on some gloves, put all the money in to a plastic bag and put it outside the front door tonight. I'm collecting all the plastic bags tonight for safety. Think of your health."
\end{quote}

\noindent
In contrast, the tweet below, which was labeled check-worthy, is spreading harmful COVID-19 misinformation which could dissuade people from getting tested.

\begin{quote}
\centering
"Coronavirus test in US is \$3,000. Here in Tokyo it’s \$50, \$166 without State ins. In much of Europe it’s free  Worse, in much of the US, it’s not even available, unreliable. And meanwhile \#POTUS recently called Corona one big “hoax.”  USA: 1st world \$\$\$, 3rd world healthcare."
\end{quote}

 We had concern that nuanced text like this may be difficult to discriminate and rank accurately.

For a journalist, the task of identifying noteworthy claims for the vetting process may be intuitive.  Their knowledge of the material, background in academic training, and experience as a journalist inform their processes and decision-making. Our learner is not coached, trained, or experienced in this area beforehand.  It receives the data and annotations provided by the annotators and learns the patterns of language to replicate their decision process.


\section{Transformer Architectures and Pre-trained Models}

\subsection{BERT}

Bidirectional Encoder Representations from Transformers (BERT) models have fundamentally changed the NLP landscape. The original BERT model's architecture consists of 12 transformers stacked on top of one another with a hidden size of 768 and 12 self-attention heads. \cite{devlin2018bert} BERT models are trained by performing unsupervised tasks, namely masked token prediction (Masked LM) and prediction of future sentences (Next Sentence Prediction) on massive amounts of data.  BERT utilizes a WordPiece tokenization scheme. \cite{schuster:2012}, and was trained on Wikipedia and the BooksCorpus \cite{zhu:2015}.  At the time of release, BERT was state-of-the-art in 11 NLP tasks.

Since initial release, many pre-trained BERT neural networks have been released.  These can be focused on new languages, or differ in size. They can be either smaller and more efficient, or larger and more comprehensive, than the original release \cite{turc:2019}. Any of these pre-trained models could serve as a base model for fine-tuning to new datasets and new tasks.

\subsection{RoBERTa}

RoBERTa, developed by Liu et al. \cite{roberta}, is an derivative of BERT which introduced modifications to the training process. The primary modifications are the provision of more training data, increasing pre-training steps with bigger batches over more data, removing Next Sentence Prediction, training on longer sequences, and dynamically changing the masking pattern applied to the training data \cite{roberta}. While RoBERTa also requires sub-word tokenization, RoBERTa uses a Byte-Pair Encoding (BPE) instead of WordPiece. \cite{sennrich:2015}  The base-roberta model was pre-trained on 160GB of text extracted from BookCorpus, English Wikipedia, CC-News, OpenWebText, and Stories (a subset of CommonCrawl Data) \cite{roberta}.

At the time of release, the RoBERTa architecture achieved state-of-the-art results on publicly available benchmark datasets such as GLUE \cite{wang:2018}, RACE \cite{lai:2017}, and SQuAD \cite{rajpurkar:2018}. Like BERT, RoBERTa models come in a variety of sizes, and choosing a model requires a trade-off between computational efficiency and model size.

While some new architectures have been released which exceed RoBERTa's performance, RoBERTa remains an accessible framework and continues to be one of the most highly ranked architectures on the SuperGLUE leaderboard.\footnote{https://super.gluebenchmark.com/leaderboard}  

\subsection{AraBERT}
AraBERT is an Arabic model developed by Wissam Antoun, Fady Baly, and Hazem Hajj at the American University of Beirut \cite{aubmindlabs}. The \textit{aubmindlab/arabert} series of models were pre-trained on Arabic documents retrieved from the web, as well as two publicly available corpora: the 1.5 billion word Arabic Corpus, and the 1 billion word Open Source International Arabic News Corpus (OSIAN).  No token count was provided for the web scraped documents. \cite{aubmindlabs}.

\subsection{ArabicBERT}
ArabicBERT is an Arabic model developed by Ali Safaya, Moutasem Abdullatif, and Deniz Yuret KUIS of Koc University. \cite{safaya:2020} ArabicBERT was trained on Wikipedia, and the OSCAR corpus \cite{ortiz:2020}, which utilized web data from CommonCrawl.  The corpus used to create the pre-trained model was, in total, 8.5 billion words.


\section{Quantitative Analysis}

\subsection{Label Balance}
The datasets for both the English and the Arabic Challenges were imbalanced.  The English Task 1 datasets contained a development dataset of 150 tweets and a training dataset of 672 tweets containing 39\% and 34\% check-worthy tweets respectively. The Arabic Task 1 training dataset provided 1,500 labeled tweets, 458 of which (31\%) were labeled check-worthy.

We will discuss provisions we make for the Arabic imbalance later in the paper.

\subsection{Vocabulary Analysis}
When utilizing pre-trained models, vocabulary used to create these models plays a critical role.  The process of fine-tuning does not allow for the addition of additional vocabulary, so these systems fallback to subword units during tokenization.  Because we were evaluating a corpus that contained emerging topics (such as COVID-19), and our pre-trained models were created at different points between 2018 and 2020, we wanted to understand what our pre-trained models contained.  We hypothesized that the models with the greatest token overlap would perform the best.

\subsubsection{English}
The token overlap between the English test dataset and RoBERTa's vocabulary file was roughly 850 tokens (54\%), with RoBERTa containing about 50K items in its vocabulary.  Many tokens missing from the RoBERTa vocabulary were related to the coronavirus topic, including several terms for COVID-19 as well as named entities, emoji, foreign languages in non-Latin script, misspellings and slang/internet chat language (LMAOOO).  No analysis was performed on the BERT vocabulary file.

\subsubsection{Arabic}The three Arabic model vocabularies contained 64K WordPieces (\textit{\\aubmindlab/bert-base-arabert}), 64K WordPieces (\textit{aubmindlab/bert-base-arabertv01}) and 32K WordPieces (\textit{asafaya/bert-base-arabic}). A rough tokenization and cleaning of the tweets in the test data set resulted in roughly 15K unique tokens. The overlap between the three Arabic model vocabulary and the Arabic test data set was roughly 8.5K tokens or 56\% of the tokens in the test data (\textit{aubmindlab/bert-base-arabertv01}), 5.5K tokens or 36\% of the tokens in the test data (\textit{asafaya/bert-base-arabic}) and 3.5K or 23\% of the tokens in the the test data (\textit{aubmindlab/bert-base-arabert}).
Some categories of vocabulary found in the test data set, but missing from the top performing model, included English words or loan words in Arabic script,  colloquial/slang, misspellings/missing spaces, named entities (names of people and places), emoji and tokens in Latin script. The \textit{asafaya/bert-base-arabic} Arabic model vocabulary also included a lot of longer WordPieces that were unlikely to be found in data. Additionally, even though the test data set contained short vowels, none of the Arabic model vocabularies had any short vowels.


\section{Approach and Results}
The datasets provided for English and Arabic contained Twitter metadata fields, but we discard these.  Our methodology only utilizes the message text of the Tweet as well as the check-worthy field containing a binary label where the positive class denoted a check-worthy claim.\footnote{We tried concatenating the text field with the pre-labeled topicID field, but this did not improve the model's performance at all, so we chose to exclude topic labels from the model.}

Competition rules required that tweets most likely to be check-worthy needed to appear at the top of each topic.  To generate rankings, we took the positive and negative class scores, generated by a sequence classification head on top the pooled output of the neural network models (whether it be BERT, RoBERTa, AraBERT, or ArabicBERT), and passed those scores through a softmax function to normalize the classification outputs. We then subtracted the negative class probability from the positive class probability. This yielded interpretable, normalized scores between 1 and -1, where higher scores reflected our model's confidence that a tweet was check-worthy. We then sorted by the difference of probabilities to produce the ranked tweets submitted to the organizers of the conference.

\subsection{English}
\subsubsection{Classification}
For our internal evaluations, we split the English training data provided into 80\% training and 20\% validation sets. We used the development set as was provided by the organizers.

We evaluated three baseline models. We fine-tuned the data over 2 epochs on the original English BERT model \cite{devlin2018bert}, a BERT model trained on COVID-19 Twitter data \cite{muller2020covid}, and the original English RoBERTa model \cite{roberta}. We assumed that the COVID-19 Twitter model would generate the highest accuracy given its deep contextual knowledge of both Twitter data and COVID-19, but of the three models, RoBERTa generated the highest precision and recall for both the positive and negative class. We chose to eliminate the previous two models and focus on optimizing RoBERTa.\footnote{In hindsight, these two should have been contributed for formal evaluation.}



In our internal evaluations, we noticed the model overfitting quickly.  To help prevent this, we added an extra mean pooling layer and dropout layer to the model. Our pooling layer takes the weights from the last layer, which were overfitting, and averages them with weights from the second-to-last layer. This reduces overfitting by smoothing out some of the weights originally calculated in the final layer. Dropout is a regularization technique that reduces overfitting by randomly omitting (or zeroing out) hidden units from the network during each training step at a probability specified by the user \cite{dropout}. By adding these two layers to the end of our RoBERTa model, we were able to improve accuracy on our test set and reduce overfitting.

After a grid search, we fine-tuned with 2 epochs, a batch size of 32, and Adam optimization with a learning rate of 1.5e-5.  The RoBERTa model was fine-tuned using the Keras API to TensorFlow.

This output was then fed through a softmax function, and the difference between the positive and negative class likelihoods were used to rank tweets within each pre-labeled topic category.

\subsubsection{Results}

Results of our fine-tuned RoBERTa model can be found in Table \ref{tab:englishresults} as \textbf{RoBERTa}.  This submission placed first place among all competing teams with a mAP of 0.8064. Our contribution narrowly beat out the second place results, which likely utilized a similar model.  We did not submit our BERT model or COVID Twitter models for formal evaluation.

\begin{table*}[]
  \caption{Accenture results from CheckThat! Task1 English.}
  \label{tab:englishresults}
\begin{tabular}{l
l llllllll}
\textbf{Entry} & \textbf{mAP}                                       & \textbf{RR}                & \textbf{R-P}               & \textbf{P@1}               & \textbf{P@3}               & \textbf{P@5}               & \textbf{P@10}              & \textbf{P@20}              & \textbf{P@30}              \\
RoBERTa     & \multicolumn{1}{r}{
0.8064} & \multicolumn{1}{r}{1.0000} & \multicolumn{1}{r}{0.7167} & \multicolumn{1}{r}{1.0000} & \multicolumn{1}{r}{1.0000} & \multicolumn{1}{r}{1.0000} & \multicolumn{1}{r}{1.0000} & \multicolumn{1}{r}{0.9500} & \multicolumn{1}{r}{0.7400}
\end{tabular}
\end{table*}

\subsection{Arabic}
\subsubsection{Classification}


For our internal evaluations, we split the Arabic training data provided into 70\% training, 20\% validation, and 10\% held-out sets. 
We evaluated four baseline Arabic BERT models retrieved from Huggingface, without any parameter tuning. \cite{Wolf2019HuggingFacesTS}. These models were \textit{Hate-speech-CNERG/ dehatebert-mono-arabic} \cite{aluru2020deep}, \textit{asafaya/bert-base-arabic} \cite{safaya:2020}, \textit{aubmindlab/bert-base-arabert} \cite{aubmindlabs}, and \textit{aubmindlab/bert-base-arabertv01} \cite{aubmindlabs}. Out of four, we found three to have promise, \textit{aubmindlab/bert-base-arabertv01}, \textit{aubmindlab/bert-base-arabert}, and \textit{asafaya/bert-base-arabic}. 



Classes were imbalanced in the Arabic training dataset with 30\% of tweets labeled as part of the check-worthy class. In order to address the imbalanced classes, we chose to upsample the positive class using machine translation via Amazon Web Services (AWS) Translate. 

Tweets from the positive class in the training and development sets were translated to English and then back to Arabic (ar$\rightarrow$en$\rightarrow$ar), appended to our training dataset, and assigned a label of check-worthy. This improved both precision and recall for check-worthy tweets, but slightly harmed the precision and recall for tweets that were not check-worthy. As the goal is to surface and rank the positive class at various levels of precision, a reduction in the F1-score of the negative class was acceptable for improving the F1-score of the positive class.

After a grid search, our final models were fine-tuned with 2 epochs, a learning rate of 2e-05, Adam optimization, and a batch size of 32. We used a Huggingface BERT sequence classification function\cite{Wolf2019HuggingFacesTS} and, like with English, added a linear layer on top of the pooled output. 


This output was then fed through a softmax function, and the difference between the positive and negative class likelihoods were used to rank tweets within each pre-labeled topic category. 

\subsubsection{Results}
Results for our Arabic evaluations can be found in Table \ref{tab:arabicresults}.
Our official submission to the competition was \textbf{AraBERT v0.1 Upsampled} and was evaluated in 1st place with a P@30 of 0.7000.  Our comparative models \textbf{AraBERT v1.0 Upsampled}\footnote{This is a rapidly evolving area of NLP.  At the time of the challenge, documentation was not yet published for AraBERT v1.0.  We did not realize v1.0 required running Farasa \cite{abdelali2016farasa} as a preprocessing step for tokenization before utilization.  We expect an Upsampled v1.0 to beat an Upsampled v0.1 when utilizing the necessary Arabic segmenter.}, \textbf{AraBERT v0.1 Unmodified}, and \textbf{ArabicBERT-Base Upsampled} were evaluated in 2nd, 3rd, and 4th place with P@30 scores of .6750, .6694, and .6639 respectively.

The benefit of back-translation to upsample the minority class can be seen by comparing \textbf{AraBERT v0.1 Upsampled} (P@30 of 0.7000) with \textbf{AraBERT v0.1 Unmodified} (P@30 of of 0.6694).  These were the same model architectures, with identical hyperparameters, but one had upsampled data, and the other did not. 


\begin{table*}[]
  \caption{Accenture results from CheckThat! Task1 Arabic}
  \label{tab:arabicresults}
\begin{tabular}{lccccc
c c}
\textbf{Entry}                      & \textbf{P@5} & \textbf{P@10} & \textbf{P@15} & \textbf{P@20} & \textbf{P@25} & \textbf{P@30} & \textbf{AP} \\
AraBERT v0.1 Upsampled                    & 0.7333       & 0.7167        & 0.7167        & 0.6875        & 0.6933        & 0.7000        & 0.6232      \\
AraBERT v1.0 Upsampled         & 0.6667       & 0.7417        & 0.7333        & 0.7125        & 0.6900        & 0.6750        & 0.5967      \\
AraBERT v0.1 Unmodified & 0.6833       & 0.7083        & 0.7111        & 0.6833        & 0.6833        & 0.6694        & 0.6035      \\
ArabicBERT-Base Upsampled         & 0.6000       & 0.6917        & 0.6944        & 0.6833        & 0.6667        & 0.6639        & 0.5947     
\end{tabular}
\end{table*}

\subsubsection{Comments: Preprocessing}
Once we had Arabic model performance baselines, we experimented with various preprocessing techniques. We assumed that these steps would reduce noise and help the Arabic BERT models better map words to tokens in its vocabulary. We performed internal evaluations involving variations of removing diacritics, stopwords, urls,  punctuation, and also of splitting underscores. We tested each of these preprocessing functions alone, as well as in combination with other preprocessing functions. We saw no increase in precision or recall from these steps. In fact, many combinations of these functions brought down our overall accuracy. We ultimately chose to forego all preprocessing.
\subsubsection{Comments: Machine Translation}
Back-translation provides the model with alternative ways to express similar concepts.  This  makes the model more robust to vocabulary not present in the training data. 

We evaluated three strategies to augment the corpus using translation data.
\begin{itemize}
\item adding back-translated data (ar$\rightarrow$en$\rightarrow$ar)
\item adding the English translation (ar$\rightarrow$en)
\item adding both the English and back-translated Arabic text (ar$\rightarrow$en; ar$\rightarrow$en$\rightarrow$ar). 
\end{itemize} We found the back-translated Arabic (without English) (ar$\rightarrow$en$\rightarrow$ar) had the provided the largest increase in accuracy on our internal evaluations. 

English was chosen as an intermediary language due solely to the fact that AWS has strong English NLP support. Future research may  explore which intermediary language translations can offer the largest performance boosts. While we may have benefited from exploring intermediary language alternatives \footnote{as well as from up-sampling the English training set}, we had to leave this for future work due to constraints in both time and budget. 

We recognize that this translation approach resulted in label leakage into the hold-out and validation sets,  resulting in overfitting on our internal evaluations. However by expanding the contextual vocabulary of the model, we had the intuition this would yield increased performance on the unseen test set.

Of all of the preprocessing and tuning steps we tried on our internal evaluations, none resulted in a larger accuracy boost than adding this back-translated data. 
\section{Future Work}
New pre-trained neural network models are being released at a rapid pace.  The trend is that they are getting larger--trained with more parameters, on larger quantities of text. Additionally, their baseline capabilities are expanding. Work like that which is presented here can be easily updated to take advantage of these new models as they become available.  The workflow a year from now will be the same, but performance will improve.  Today, BERT and similar pre-trained models have become the new baseline.  These systems yield fantastic results, with little training data required for fine-tuning.  

As larger models are created and released, the models become more difficult to understand.  Classification and ranking is helpful to support SMEs performing their work, but full decision support systems cannot be black boxes, and need to be able to explain why they made the suggestions they did.  We are working on improving the explainability of these models to provide better support to decision makers. 

\section{Conclusions}

This paper introduced work by Accenture on using BERT and RoBERTa models to classify and rank unsubstantiated claims in social media for professional fact-checking.  We demonstrate 5 models.   We submitted one model to the English track, and placed 1st with a mAP of .8064.  We submitted 4 models to the Arabic track, yielding 1st (P@30=.7000), 2nd (P@30=.6750), 3rd (P@30=.6694), and 4th (P@30=.6639) place.


\begin{thebibliography}{10}
\providecommand{\url}[1]{\texttt{#1}}
\providecommand{\urlprefix}{URL }
\providecommand{\doi}[1]{https://doi.org/#1}

\bibitem{abdelali2016farasa}
Abdelali, A., Darwish, K., Durrani, N., Mubarak, H.: Farasa: A fast and furious
  segmenter for arabic. In: Proceedings of the 2016 conference of the North
  American chapter of the association for computational linguistics:
  Demonstrations. pp. 11--16 (2016)

\bibitem{aluru2020deep}
Aluru, S.S., Mathew, B., Saha, P., Mukherjee, A.: Deep learning models for
  multilingual hate speech detection. arXiv preprint arXiv:2004.06465  (2020)

\bibitem{aubmindlabs}
Antoun, W., Baly, F., Hazem, H.: {AraBERT}: Transformer-based model for arabic
  language understanding. In: Proceedings of the 4th Workshop on Open-Source
  Arabic Corpora and Processing Tools, with a Shared Task on Offensive Language
  Detection. pp. 9--15 (2020), \url{https://arxiv.org/pdf/2003.00104v2.pdf}

\bibitem{CLEF_LNCS:20}
Arampatzis, A., Kanoulas, E., Tsikrika, T., Vrochidis, S., Joho, H., Lioma, C.,
  Eickhoff, C., Névéol, A., Cappellato, L., Ferro, N. (eds.): Experimental IR
  Meets Multilinguality, Multimodality, and Interaction Proceedings of the
  Eleventh International Conference of the CLEF Association (CLEF 2020). LNCS
  (12260), Springer (2020)

\bibitem{clef-checkthat-lncs:2020}
Barr\'{o}n-Cede{\~n}o, A., Elsayed, T., Nakov, P., {Da San Martino}, G.,
  Hasanain, M., Suwaileh, R., Haouari, F., Babulkov, N., Hamdan, B., Nikolov,
  A., Shaar, S., {Sheikh Ali}, Z.: {Overview of CheckThat! 2020}: Automatic
  identification and verification of claims in social media. In: Arampatzis
  et~al.  \cite{CLEF_LNCS:20}

\bibitem{brown:2020}
Brown, T.B., Mann, B., Ryder, N., Subbiah, M., Kaplan, J., Dhariwal, P.,
  Neelakantan, A., Shyam, P., Sastry, G., Askell, A., Agarwal, S.,
  Herbert-Voss, A., Krueger, G., Henighan, T., Child, R., Ramesh, A., Ziegler,
  D.M., Wu, J., Winter, C., Hesse, C., Chen, M., Sigler, E., Litwin, M., Gray,
  S., Chess, B., Clark, J., Berner, C., McCandlish, S., Radford, A., Sutskever,
  I., Amodei, D.: Language models are few-shot learners (2020)

\bibitem{clef2020-workingnotes}
Cappellato, L., Eickhoff, C., Ferro, N., Névéol, A. (eds.): Working Notes of
  CLEF 2020---Conference and Labs of the Evaluation Forum (2020)

\bibitem{checkthat2020}
Committee, O.: Tasks 1 \& 5: Check-worthiness,
  \url{https://sites.google.com/view/clef2020-checkthat/tasks/tasks-1-5-check-worthiness}

\bibitem{devlin2018bert}
Devlin, J., Chang, M.W., Lee, K., Toutanova, K.: {BERT}: Pre-training of deep
  bidirectional transformers for language understanding. arXiv preprint
  arXiv:1810.04805  (2018)

\bibitem{clef-checkthat-ar:2020}
Hasanain, M., Haouari, F., Suwaileh, R., Ali, Z., Hamdan, B., Elsayed, T.,
  Barr\'{o}n-Cede{\~n}o, A., {Da San Martino}, G., Nakov, P.: Overview of
  {CheckThat!} 2020 {A}rabic: Automatic identification and verification of
  claims in social media. In: Cappellato et~al.  \cite{clef2020-workingnotes}

\bibitem{dropout}
Hinton, G.E., Srivastava, N., Krizhevsky, A., Sutskever, I., Salakhutdinov,
  R.R.: Improving neural networks by preventing co-adaptation of feature
  detectors. arXiv preprint arXiv:1207.0580  (2012)

\bibitem{howard:2018}
Howard, J., Ruder, S.: Fine-tuned language models for text classification. CoRR
   \textbf{abs/1801.06146} (2018), \url{http://arxiv.org/abs/1801.06146}

\bibitem{lai:2017}
Lai, G., Xie, Q., Liu, H., Yang, Y., Hovy, E.: Race: Large-scale reading
  comprehension dataset from examinations (2017)

\bibitem{roberta}
Liu, Y., Ott, M., Goyal, N., Du, J., Joshi, M., Chen, D., Levy, O., Lewis, M.,
  Zettlemoyer, L., Stoyanov, V.: {RoBERTa}: {A} robustly optimized {BERT}
  pretraining approach. CoRR  \textbf{abs/1907.11692} (2019),
  \url{http://arxiv.org/abs/1907.11692}

\bibitem{muller2020covid}
M{\"u}ller, M., Salath{\'e}, M., Kummervold, P.E.: Covid-twitter-bert: A
  natural language processing model to analyse {COVID-19} content on twitter.
  arXiv preprint arXiv:2005.07503  (2020)

\bibitem{ortiz:2020}
Ortiz~Suárez, P.J., Romary, L., Sagot, B.: A monolingual approach to
  contextualized word embeddings for mid-resource languages. Proceedings of the
  58th Annual Meeting of the Association for Computational Linguistics  (2020).
  \doi{10.18653/v1/2020.acl-main.156},
  \url{http://dx.doi.org/10.18653/v1/2020.acl-main.156}

\bibitem{peters:2018}
Peters, M.E., Neumann, M., Iyyer, M., Gardner, M., Clark, C., Lee, K.,
  Zettlemoyer, L.: Deep contextualized word representations. In: Proc. of NAACL
  (2018)

\bibitem{radford:2019}
Radford, A., Wu, J., Child, R., Luan, D., Amodei, D., Sutskever, I.: Language
  models are unsupervised multitask learners. Tech. rep., OpenAI, San
  Francisco, CA, USA (2019)

\bibitem{rajpurkar:2018}
Rajpurkar, P., Jia, R., Liang, P.: Know what you don't know: Unanswerable
  questions for squad (2018)

\bibitem{rasmy2020medbert}
Rasmy, L., Xiang, Y., Xie, Z., Tao, C., Zhi, D.: Med-bert: pre-trained
  contextualized embeddings on large-scale structured electronic health records
  for disease prediction (2020)

\bibitem{safaya:2020}
Safaya, A., Abdullatif, M., Yuret, D.: Kuisail at semeval-2020 task 12:
  Bert-cnn for offensive speech identification in social media. In: Proceedings
  of the International Workshop on Semantic Evaluation (SemEval) (2020)

\bibitem{schuster:2012}
Schuster, M., Nakajima, K.: Japanese and {K}orean voice search. In: 2012 IEEE
  International Conference on Acoustics, Speech and Signal Processing (ICASSP).
  pp. 5149--5152. IEEE (2012)

\bibitem{sennrich:2015}
Sennrich, R., Haddow, B., Birch, A.: Neural machine translation of rare words
  with subword units (2015)

\bibitem{clef-checkthat-en:2020}
Shaar, S., Nikolov, A., Babulkov, N., Alam, F., Barr\'{o}n-Cede{\~n}o, A.,
  Elsayed, T., Hasanain, M., Suwaileh, R., Haouari, F., {Da San Martino}, G.,
  Nakov, P.: Overview of {CheckThat!} 2020 {E}nglish: Automatic identification
  and verification of claims in social media. In: Cappellato et~al.
  \cite{clef2020-workingnotes}

\bibitem{sharir:2020}
Sharir, O., Peleg, B., Shoham, Y.: The cost of training {NLP} models: A concise
  overview. arXiv preprint arXiv:2004.08900v1  (2020)

\bibitem{sun:2019}
Sun, Y., Wang, S., Li, Y., Feng, S., Chen, X., Zhang, H., Tian, X., Zhu, D.,
  Tian, H., Wu, H.: Ernie: Enhanced representation through knowledge
  integration. arXiv preprint arXiv:1904.09223  (2019)

\bibitem{turc:2019}
Turc, I., Chang, M.W., Lee, K., Toutanova, K.: Well-read students learn better:
  On the importance of pre-training compact models (2019)

\bibitem{wang:2018}
Wang, A., Singh, A., Michael, J., Hill, F., Levy, O., Bowman, S.R.: Glue: A
  multi-task benchmark and analysis platform for natural language understanding
  (2018)

\bibitem{Wolf2019HuggingFacesTS}
Wolf, T., Debut, L., Sanh, V., Chaumond, J., Delangue, C., Moi, A., Cistac, P.,
  Rault, T., Louf, R., Funtowicz, M., Brew, J.: Huggingface's transformers:
  State-of-the-art natural language processing. ArXiv  \textbf{abs/1910.03771}
  (2019)

\bibitem{zhu:2015}
Zhu, Y., Kiros, R., Zemel, R.S., Salakhutdinov, R., Urtasun, R., Torralba, A.,
  Fidler, S.: Aligning books and movies: Towards story-like visual explanations
  by watching movies and reading books. CoRR  \textbf{abs/1506.06724} (2015),
  \url{http://arxiv.org/abs/1506.06724}

\end{thebibliography}
\end{document}